\documentclass[letterpaper]{article} 
\usepackage[preprint]{aaai2027}  
\usepackage[hyphens]{url}  
\usepackage{graphicx} 
\usepackage{natbib}  
\usepackage{caption} 
\usepackage{algorithm}
\usepackage{algorithmic}
\usepackage{amsfonts}

\usepackage{newfloat}
\usepackage{listings}
\DeclareCaptionStyle{ruled}{labelfont=normalfont,labelsep=colon,strut=off} 
\floatstyle{ruled}
\newfloat{listing}{tb}{l
st}{}
\floatname{listing}{Listing}

\usepackage{booktabs}
\usepackage{amsmath}

\usepackage{tabularx}
\usepackage{multirow}

\newcolumntype{C}{>{\centering\arraybackslash}X}

\usepackage{pifont}

\title{Counting Beyond Instances: BunchCount for Group-Individual Object Counting}
\author{
    Rui Wang\textsuperscript{\rm 1},
    Junyi Huang\textsuperscript{\rm 1},
    Jiahui Li\textsuperscript{\rm 1},
    Qiao Yu\textsuperscript{\rm 2},
    Yixue Hao\textsuperscript{\rm 1},
    Long Hu\textsuperscript{\rm 1}\corresponding,
    Baoru Huang\textsuperscript{\rm 3}
}
\affiliations{
    \textsuperscript{\rm 1}School of Computer Science and Technology,
    Huazhong University of Science and Technology, Wuhan, China\\
    \textsuperscript{\rm 2}Shanghai Artificial Intelligence Laboratory, China\\
    \textsuperscript{\rm 3}Department of Computer Science,
    University of Liverpool, Liverpool, United Kingdom\\
    hulong@hust.edu.cn
}

\begin{document}

\maketitle

\begin{abstract}

Visual counting is commonly formulated at the instance level, aiming to estimate how many objects of a queried category appear in an image. 
However, real-world counting often involves higher-level semantic units formed by multiple instances, such as a bunch of grapes, a stack of plates, or a pair of shoes.
This exposes a key limitation of existing counting formulations, which mainly focus on \emph{what to count}, while largely overlooking \emph{at which semantic unit to count}. We introduce \textbf{Group-Individual Object Counting} (GIC), a new setting that requires models to count both individual objects and semantic groups within a unified framework. To support this new task, we present \textbf{BunchCount}, a real-world benchmark with 1,330 images, 89,254 individual annotations, and 11,065 group annotations.
BunchCount provides paired individual-group annotations within the same image and explicitly records containment relations between each group and its constituent individuals.
Experiments on BunchCount show that current advanced counting models perform well on individual instances but fail to count semantic groups more accurately.
To mitigate semantic granularity conflict, we propose a counting-unit guided relational counting framework, which exploits group-individual containment relations to regularize cross-granularity representations during training. 
Our method substantially improves group-level counting while better preserving individual-level counting ability, establishing a strong baseline for counting beyond instances.

\end{abstract}

\begin{links}
    \link{Code}{https://github.com/deciiimal/BunchCount}
\end{links}

\section{Introduction}

\begin{figure}[!t]
\centering
{\small
\begin{tabular}{@{}c@{}c@{}}
  \begin{tabular}[c]{@{}c@{}}
    \includegraphics[width=0.47\columnwidth]{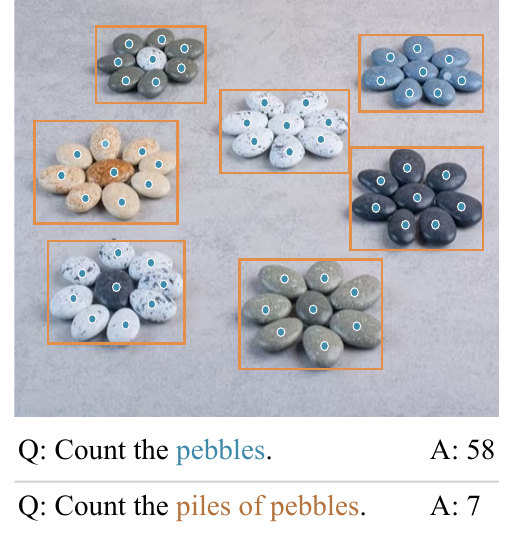}
  \end{tabular}
  &
  \begin{tabular}[c]{@{}c@{}}
    \includegraphics[width=0.52\columnwidth]{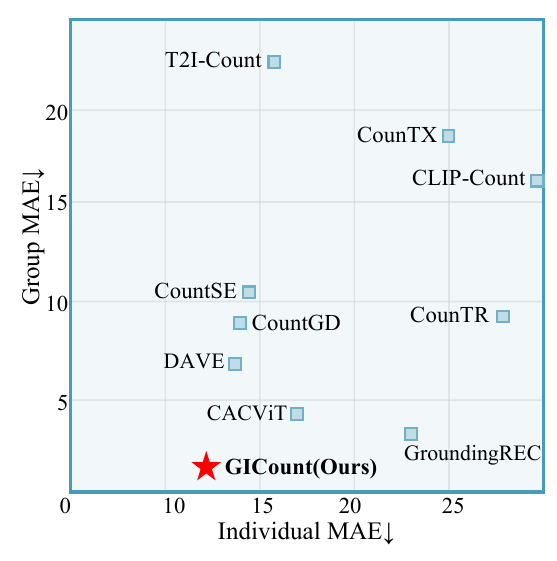}
  \end{tabular}
  \\
  \makebox[0.47\columnwidth][c]{%
    \shortstack[l]{%
      (a) GIC task example
    }
  }
  &
  \makebox[0.52\columnwidth][c]{%
    \shortstack[l]{%
      (b) Performance comparisons
    }
  }
\end{tabular}
}
\caption{Illustration of group-individual object counting. (a) Group-individual object counting. (b) Counting model performance comparisons on the BunchCount. }
\label{fig:introduction}
\end{figure}

Object counting aims to estimate the number of instances of target objects in an image \cite{Ranjan_2021_CVPR}.
Recent studies have extended counting from fixed categories to more flexible target specifications, such as visual exemplars \cite{Ranjan_2021_CVPR, liu2022countr} or text descriptions \cite{Amini-Naieni_2023_BMVC, 10.1145/3581783.3611789}. 
Despite these advances, most existing counting paradigms still treat individual instances as the basic counting units.
They have substantially expanded \emph{what to count}, but rarely examine \emph{which semantic unit} should be counted.

In many practical scenarios, object counting often involves beyond individual instances for the same set of objects, which can be counted either individually or according to the semantic groups they form.
For example, grapes may be counted as individual grapes or as bunches, shoes as individual shoes or as pairs, and plates as individual plates or as stacks. 
These two levels of counting are both meaningful and serve different practical needs \cite{PEREZZAVALA2018136}. 
Motivated by this observation, we introduce a new counting paradigm, {\textbf{Group-Individual Object Counting}} \textbf{(GIC)}, as shown in Figure~\ref{fig:introduction}(a). 
Given an image and a prompt, a model is required to count either individual objects or the semantic groups, according to the counting unit specified by the prompt.

The GIC task is distinct from existing counting formulations. 
Class-agnostic \cite{Ranjan_2021_CVPR} and fine-grained counting \cite{Dai_2024_CVPR, 11302471} extend the range of countable categories or attributes, but still typically regard individual instances as the basic counting units. 
Related vision tasks have explored group structure. Group-aware perception \cite{Zhang_2016_CVPR} aims to recover group structures or uses groups as intermediate regions, while Locount \cite{Cai_Wen_Zhang_Du_Wang_2021} localizes regions of heavily overlapping same-category objects and estimates the number of constituent instances within each region.
Under this background, \textbf{GIC formulates counting-unit selection as an explicit component of object counting}, allowing the prompt to specify either individual objects or the semantic groups they form.
However, there is no suitable benchmark to study this problem.
Existing counting datasets usually annotate targets at a single granularity \cite{Ranjan_2021_CVPR}, and therefore cannot evaluate whether a model can switch between individual-level and group-level counting for the same category in the same image. 
Moreover, since each semantic group is composed of specific constituent instances, group or individual annotations alone are insufficient to fully characterize the task. 
To fill this gap, we construct \textbf{BunchCount}, a real-world benchmark containing 1,330 images, 89,254 individual annotations, and 11,065 group annotations. 
Each image provides paired individual and group annotations, together with explicit group-individual containment relations that capture real-world group structures, including structural arrangement, physical connection, fixed cardinality, and spatial aggregation.

Furthermore, we examine whether existing counting models can serve as direct solutions to the new GIC task.
Representative visual-exemplar~\cite{liu2022countr}, text-guided~\cite{Amini-Naieni_2023_BMVC}, and grounding-based counters~\cite{NEURIPS2024_57c56985} can count repeated target instances, making them natural baselines for GIC.
However, their established individual-counting capability does not reliably transfer to semantic groups, as illustrated in Figure~\ref{fig:introduction}(b).
Naive joint fine-tuning further exposes a semantic granularity conflict that improving group counting may degrade individual counting. 
It derives from the fact that a group is a relational unit formed by its group structures rather than a fixed-appearance object. Thus, simply mixing annotations from both granularities is insufficient to learn a unified counter.
To address this challenge, we propose \textbf{GICount}, a counting-unit guided relational counting framework. GICount models the queried target as a counting unit, and exploits group-individual containment relations to regularize cross-granularity representations. This enables the model to count the intended individual or group unit while preserving structural consistency across counting granularities.
Our main contributions are summarized as follows:
\begin{itemize}
    \item We introduce GIC, a new counting task that extends object counting from target-category selection to semantic counting-unit selection, requiring a model to count either individual objects or the semantic groups.
    \item We construct BunchCount, a real-world benchmark for GIC, containing paired individual- and group-level annotations in the same images, together with explicit containment relations between the group and its instances.
    \item We conduct a systematic benchmark study on BunchCount and show that individual-counting capability does not naturally transfer to group counting, revealing semantic granularity conflict across counting units.
    \item We propose GICount, a counting-unit guided relational counting framework. It models group-unit prompt and exploits group-individual containment relations to improve group counting while preserving individual counting.

\end{itemize}

\section{Related Work}

\begin{figure*}[!t]
  \centering
  \includegraphics[width=0.99\textwidth]{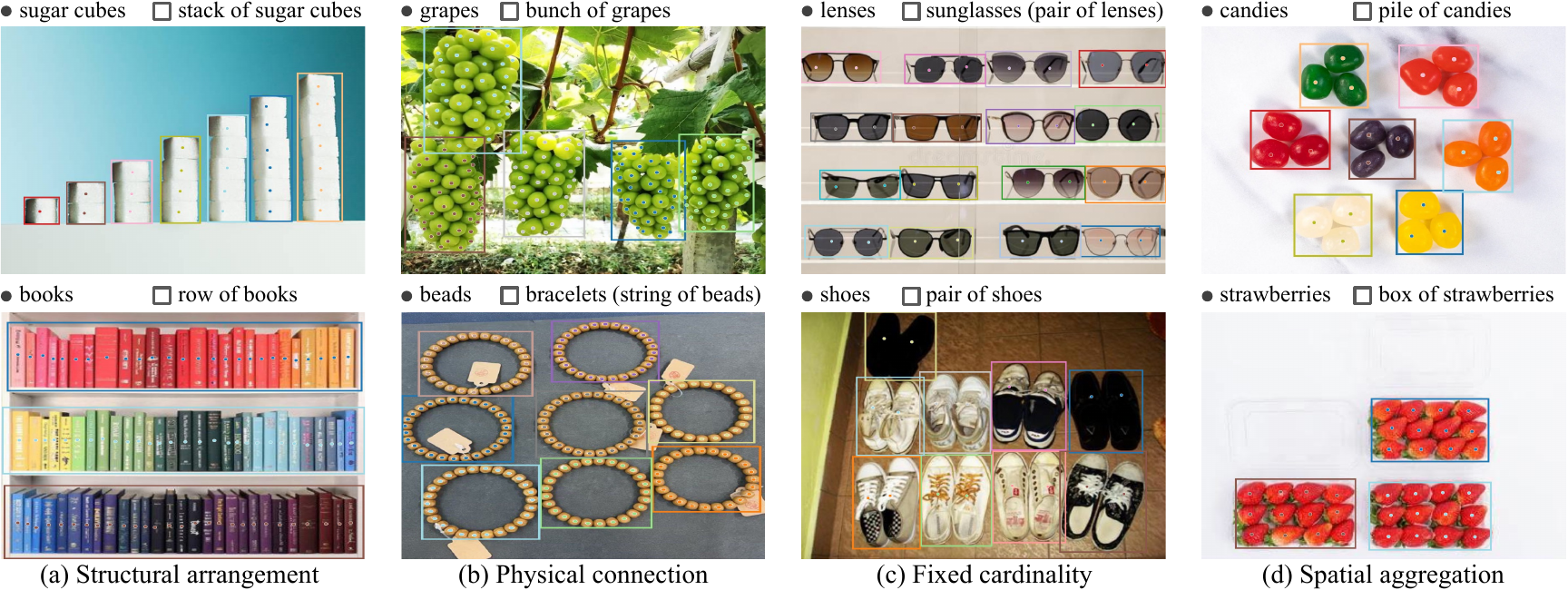}
  \caption{Representative samples from BunchCount, covering (a) structural arrangement, (b) physical connection, (c) fixed cardinality, and (d) spatial aggregation. Dots and boxes indicate individual and group annotations, respectively.}
  \label{fig:dataset-samples}
\end{figure*}

\begin{figure*}[!t]
\centering
\begin{tabular}{
  @{}c
  @{}c
  @{\hspace{0.03\textwidth}}c@{}
}

\begin{tabular}[c]{@{}c@{}}
  \includegraphics[width=0.32\textwidth]
    {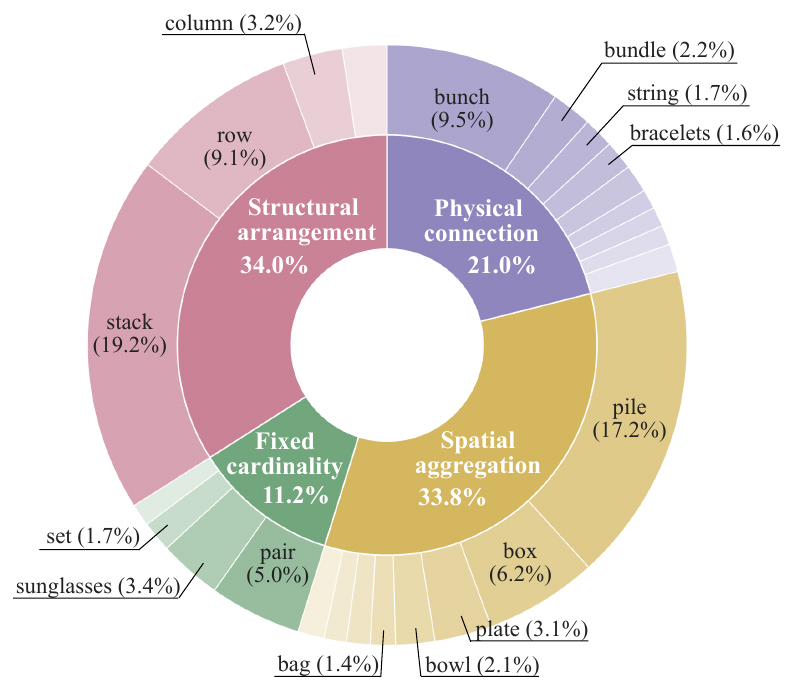}
\end{tabular}

&

\begin{tabular}[c]{@{}c@{}}
  \includegraphics[width=0.23\textwidth]
    {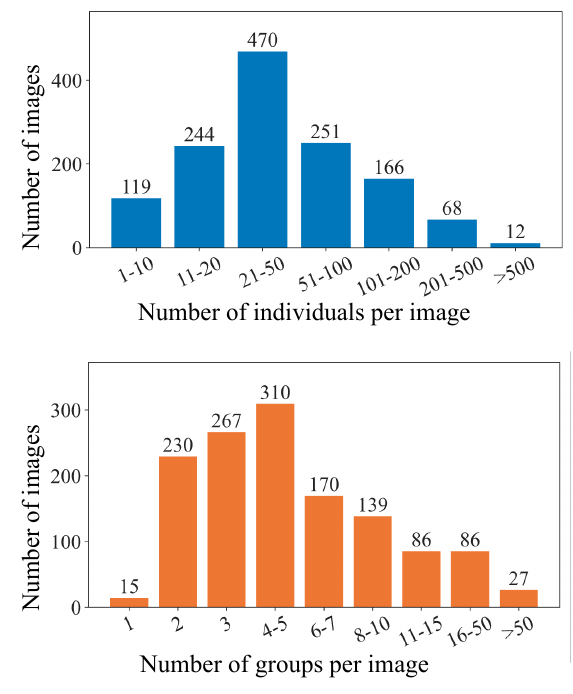}
\end{tabular}

&

\begin{tabular}[c]{@{}c@{}}
  {
  \small
  \setlength{\tabcolsep}{2pt}
  \begin{tabularx}{0.4\textwidth}{@{}lCCCC@{}}
    \toprule
    \multirow{2}{*}{Dataset} &
    \multirow{2}{*}{Images} &
    \multirow{2}{*}{Classes} &
    \multicolumn{2}{c}{Annotations} \\
    \cmidrule(lr){4-5}
    & & & Ind. avg. & Grp. avg. \\
    \midrule
    UCF\_CC\_50~\shortcite{Idrees_2013_CVPR} & 50  & 1  & 1279 & -- \\
    CountBench~\shortcite{Paiss_2023_ICCV}  & 540 & -- & 6 & --  \\
    PairTally~\shortcite{11302471}   & 681  & 54  & 160 & --\\
    CARPK~\shortcite{Hsieh_2017_ICCV}      & 1448& 1  & 62 & -- \\
    FSC-147~\shortcite{Ranjan_2021_CVPR}     & 6135& 147 & 56 & --  \\
    REC-8k~\shortcite{Dai_2024_CVPR}      & 8011 & --& 36 & --  \\
    \midrule
    \textbf{BunchCount~(Ours)}  & \textbf{1330} & \textbf{86} & \textbf{67} & \textbf{8} \\
    \bottomrule
  \end{tabularx}
  }
  \\
  \addlinespace[8pt]
  \makebox[0.4\textwidth][c]{%
    \shortstack[l]{%
      (c) Comparison with popular counting datasets
    }
  }
\end{tabular}
\label{tab:dataset-comparison}

\\

\makebox[0.32\textwidth][c]{%
  \shortstack[l]{%
    (a) Distribution of group structures\\
    in BunchCount
  }
}
&
\makebox[0.23\textwidth][c]{%
  \shortstack[l]{%
    (b) Number of individual and\\ 
    group counts per image
  }
}
&
{}

\end{tabular}

\caption{Statistics and comparison of BunchCount.
(a) Distribution of group structure types and descriptions.
(b) Distributions of the numbers of individual and group annotations
per image.
(c) Comparison with existing counting datasets.}
\label{fig:dataset-statistics}
\end{figure*}

\subsection{Object Counting Benchmarks}
Early counting benchmarks are largely developed for specific domains, such as crowds~\cite{Zhang_2016_CVPR_MCNN, Sindagi_2019_ICCV, 9153156}, cars~\cite{Hsieh_2017_ICCV} and cells~\cite{NIPS2010_fe73f687}, with predefined target categories. FSC-147~\cite{Ranjan_2021_CVPR} establishes the widely adopted benchmark for class-agnostic counting, where models estimate the number of objects from a target category. Subsequent benchmarks broaden this paradigm through target localization~\cite{10.1007/978-3-031-20044-1_20}, multi-category annotation~\cite{mondal2025omnicount}, and counting question answering~\cite{Deitke_2025_CVPR}, expanding both the range of countable targets and the ways in which they can be specified.

More recent benchmarks provide finer control over target selection. REC-8k~\cite{Dai_2024_CVPR} uses referring expressions to select instances according to attributes and contextual cues, while PairTally~\cite{11302471} evaluates counting in the presence of visually or semantically similar distractors. KubriCount~\cite{liu2026kubricount} further supports target descriptions at multiple semantic abstraction levels. These benchmarks refine the semantics of a counting query and determine which object instances should be counted, but still treat each selected instance as the implicit counting unit. BunchCount addresses this missing dimension by allowing objects of the same category within an image to be counted either as individual objects or the semantic groups formed by multiple instances. Its paired individual and group annotations, together with their containment relations, enable counting units to be studied while controlling for visual content and target category.

\subsection{Group Understanding and Relation Modeling}

Group-level visual understanding studies how multiple people can be aggregated into a coherent whole for collective activity recognition and social group analysis~\cite{Ibrahim_2016_CVPR,Gavrilyuk_2020_CVPR,Li_2021_ICCV,Kim_2022_CVPR}. Related application~\cite{Cai_Wen_Zhang_Du_Wang_2021} has also represented clusters of same-category objects as localized regions and estimated the number of instances they contain. 
In relational modeling, Relationformer~\cite{10.1007/978-3-031-19836-6_24} introduces a dedicated relation token to model relations among DETR-style object queries, within a broader line of transformer-based scene graph methods~\cite{Im_2024_CVPR_EGTR,10105507,Li_2022_CVPR,Hayder_2024_CVPR}. Together, these works provide complementary precedents for aggregating multiple instances and reasoning over their relations. 
In object counting, GIC introduces counting granularity as an explicit query dimension, treating individuals and semantic groups as alternative counting units, with BunchCount supporting the task through paired individual and group annotations in each image and containment relations linking groups to their constituent instances.

\section{Group--Individual Object Counting}

\begin{figure*}[!t]
  \centering
  \includegraphics[width=\textwidth]{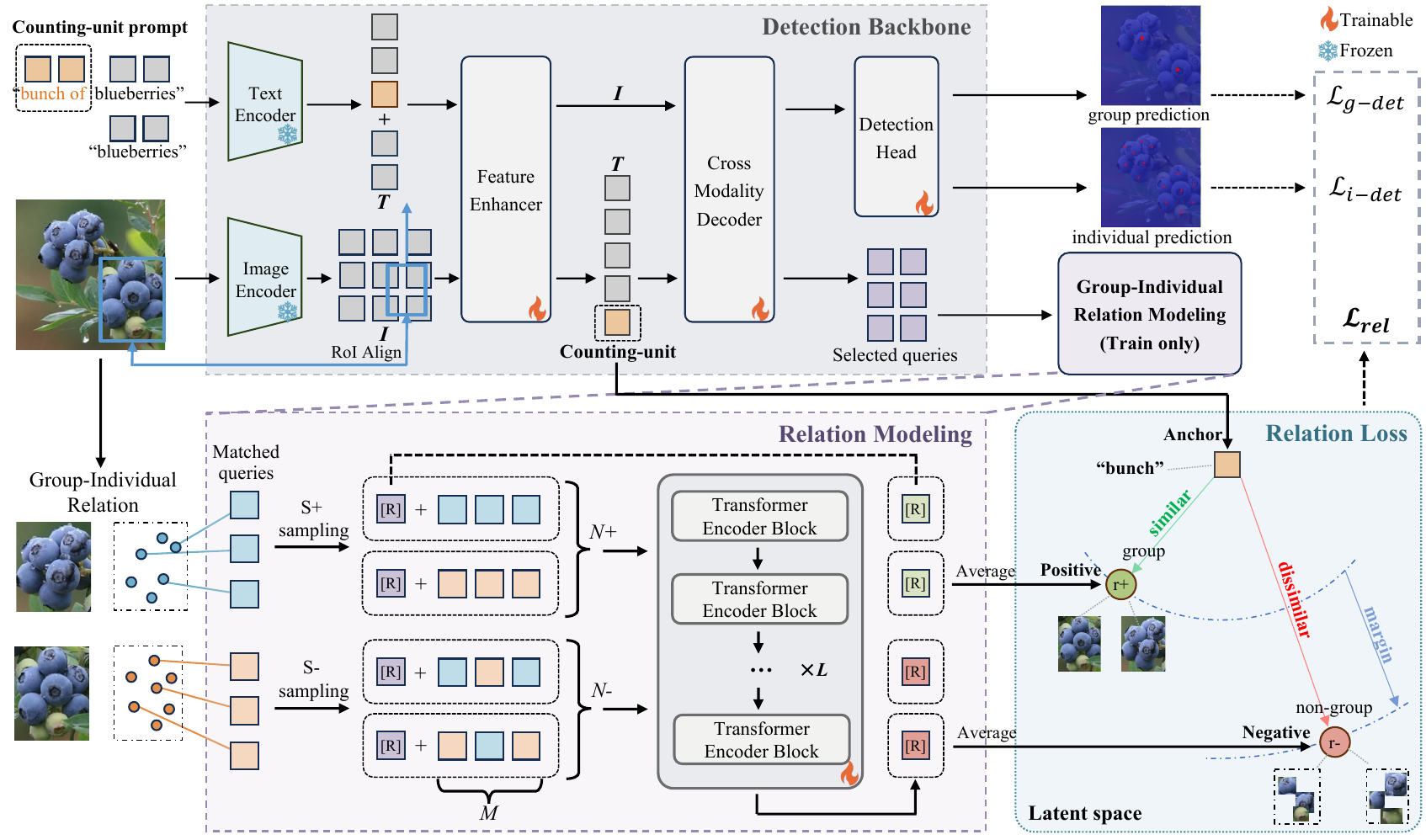}
  \caption{Overview of the GICount framework. A shared detection backbone receives a counting-unit prompt and predicts the group or individual counts, while the training-only group-individual relation modeling aggregates matched queries.}
  \label{fig:method}
\end{figure*}

Object counting is typically defined at the individual-instance level. 
We extend it to Group-Individual Object Counting (GIC), where the prompt specifies both the target category and the semantic counting unit.
Given an image $I$ and a target category $c$, let $\mathcal{I}_c=\{i_1,\ldots,i_{N_{\mathrm{ind}}}\}$ denote the set of individual instances. A semantic group is a countable higher-level unit formed by multiple instances through connection, aggregation, arrangement, or fixed cardinality. We denote the group set as $\mathcal{G}_c=\{g_1,\ldots,g_{N_{\mathrm{grp}}}\}$. Each group $g_j$ is associated with its constituent instances under a group-individual containment relation. Each individual belongs to at most one group, while some individuals may remain ungrouped.

Given a prompt $P$, the target category is $c$ and the specified counting unit is $u\in\{\mathrm{ind},\mathrm{grp}\}$. The target count $N_u$ is 
$$
N_u=
\begin{cases}
|\mathcal{I}_c|, & u=\mathrm{ind},\\
|\mathcal{G}_c|, & u=\mathrm{grp}.
\end{cases}
$$
Thus, the same image and category may have different valid counts under different prompts, e.g., counting individual grapes versus counting bunches of grapes. GIC requires the model to follow the specified semantic unit and count either individual instances or semantic groups accordingly.

\section{BunchCount Dataset}

To support GIC, we construct BunchCount, a real-world benchmark that provides paired individual- and group-level annotations within the same image.

\subsection{Data Collection and Annotation}

We collect images from FSC-147~\cite{Ranjan_2021_CVPR}, Locount~\cite{Cai_Wen_Zhang_Du_Wang_2021}, COCO~\cite{10.1007/978-3-319-10602-1_48}, OpenImages~\cite{OpenImages}, and the web, and re-annotate them under a unified GIC protocol. BunchCount focuses on real-world scenes involving structural arrangement, physical connection, fixed cardinality and spatial aggregation as shown in Figure~\ref{fig:dataset-samples}, while retaining only images with identifiable instances, clear group boundaries, and reliable group relations. Each image is annotated at two semantic units: individual objects are labeled by center points, and semantic groups are labeled by bounding boxes enclosing their constituent instances. 
We further record group-individual containment relations by linking each group box to its member points, also allowing explicitly ungrouped individuals. 
For prompt-based counting, we additionally provide visual exemplars, including three individual boxes and one to three group boxes per image. The annotation was conducted by six annotators over approximately 183 hours, followed by three rounds of correction and consistency checking.

\subsection{Dataset Statistics and Split}
BunchCount contains {1,330} real-world images, {89,254} individual point annotations, and {11,065} group bounding-box annotations. Each image includes 4--941 individuals and 1--253 groups, with averages of 67 individuals and 8 groups, respectively. The dataset covers diverse group structures and count distributions shown in Figure~\ref{fig:dataset-statistics}(a) and Figure~\ref{fig:dataset-statistics}(b), which is compared with representative counting benchmarks such as UCF\_CC\_50~\cite{Idrees_2013_CVPR}, FSC-147~\cite{Ranjan_2021_CVPR} and REC-8k~\cite{Dai_2024_CVPR} in Figure~\ref{fig:dataset-statistics}(c). Furthermore, we split BunchCount by target category into mutually disjoint training, validation, and test sets, containing 63, 9, and 14 categories with 869, 151, and 310 images, respectively. For categories overlapping with FSC-147, we follow its original split. 
Each image provides paired individual- and group-level counting samples. Group-structure descriptions are shared across splits evaluate whether models can transfer learned group-structure semantics to unseen object categories. Additional dataset analysis are provided in Appendix~A.

\section{GICount Framework}

We present GICount, a counting-unit guided relational counting framework for GIC in Figure~\ref{fig:method}. The framework contains two key components: the counting-unit prompt and group-individual relation modeling.

\subsection{Overview}

GICount follows a detection-based counting architecture built on GroundingDINO~\cite{10.1007/978-3-031-72970-6_3}, a vision-language detector that localizes open-vocabulary targets from textual prompts. This architecture is suitable for GIC because the model localizes and enumerates the semantic counting unit specified by the prompt, rather than regressing a global count from local repetitions. 
As shown in Figure~\ref{fig:method}, given an $I\in \mathbb{R}^{H\times W\times 3}$, a textual prompt $t$, and visual exemplars, GICount first encodes visual and textual features with the GroundingDINO backbone and then decodes target regions corresponding to the queried unit.

To adapt this architecture to the GIC task, GICount introduces two key components. First, counting-unit prompt distinguishes category semantics from group-structure semantics, enabling the model to represent either individual instances or semantic groups as counting units.
Second, group-individual relation modeling uses containment relations to align group representations with the aggregated representations of their constituent individuals during training. The final count is obtained by enumerating localized predictions, allowing GICount to improve group-level counting while preserving individual-level counting capability.

\subsection{Counting-unit Prompt} 

In the GIC task,  the counting-unit prompt specifies both the target category and the semantic counting unit. For individual-level queries, the unit is the object instance itself, and the prompt mainly encodes category semantics, e.g., ``apples''. 
For group-level queries, the unit is a semantic group, whose prompt contains a category component describing the constituent object type and a group-structure component describing how multiple instances form a countable whole, such as aggregation, arrangement, physical connection, or fixed cardinality. 
For explicit group prompts,  we separate it from the category term, e.g., ``pile of'' and ``apples'' in ``pile of apples''. 
For holistic group names with implicit structure, e.g., ``trains'' composed of multiple carriages, we insert a learnable group-unit token to represent the latent group structure. 
The complete prompt is used for text encoding, cross-modal fusion, and decoding, while the group-unit representation is extracted for subsequent group-individual relation modeling during training. 
This design enables GICount to ground the queried target as a semantic counting unit.

\subsection{Group--Individual Relation Modeling}

Group-level counting requires the model to associate the textual group-unit semantics with the visual organization among individuals. For example, in the prompt ``pile of apples'', the relation among apples belonging to the same pile should correspond to the structural semantics of ``pile,'' whereas an arbitrary combination of apples drawn from different piles should not express the same group structure.
To impose this distinction, we construct positive and negative sets of individual queries, aggregate the relations within each set, and employ metric learning to align textual group-structure semantics with visual relations among individuals.

\noindent\textbf{Positive and Negative Relation-Set Sampling.} For a paired group-level sample and individual-level sample from the same image, we obtain the group-structure semantic representation $s$ from the counting-unit prompt. 
In the individual branch, predicted points are matched to ground-truth individual points through bipartite matching, and the decoder queries are denoted as $\{q_i\}$. 
According to the annotated containment relations, each matched query is assigned to its associated group if it belongs to one. We then sample positive relation sets from individuals within the same group. Negative relation sets are constructed by mixing individuals from different groups and/or ungrouped individuals. For each paired sample, we sample $N^{+}$ positive sets and $N^{-}$ negative sets, each containing at most $M$ individual queries. When fewer than $M$ queries are available, the remaining positions are padded with padding masks.

\noindent\textbf{Relation Aggregation.}
To encode the organization of each sampled individual set, we prepend a learnable $[\mathrm{REL}]$ token to its query sequence and feed the sequence into a Transformer-based relation aggregator with $L$ blocks. 
The encoded $[\mathrm{REL}]$ token is used as the relation representation of the corresponding set, yielding positive set representations $r_i^{+}$ and negative set representations $r_j^{-}$. 
Finally, mean pooling is applied separately over the $N^{+}$ positive and the $N^{-}$ negative representations to obtain the sample-level positive and negative relation representations $r^{+}$ and $r^{-}$.

\noindent\textbf{Relation Loss.}
We use cosine distance $d(\cdot,\cdot)$ to align the group-unit representation with valid individual-set relations. For the $k$-th paired sample, we compute the positive distance $d(s_k,r_k^{+})$ and negative distance $d(s_k,r_k^{-})$. 
The relation objective is defined as
\begin{equation}
\begin{split}
\mathcal{L}_{\mathrm{rel}}
=\frac{1}{K}\sum_{k=1}^{K}
[&\max(0,d(s_k,r_k^{+})-d(s_k,r_k^{-})+m) \\
 & + d(s_k, r^+_k)],
\end{split}
\end{equation}
where $s$ is the semantic anchor, $K$ denotes the number of paired samples and $m$ is the triplet margin. As illustrated in Figure~\ref{fig:method}, the objective pulls together the anchor and positive relations formed by individuals from the same group, while pushing apart the anchor and negative relations constructed from different groups. This supervision establishes an explicit correspondence between textual group-structure semantics and visual patterns of individual organization.

\begin{table*}[!t]
\centering
{\small
\setlength{\tabcolsep}{4pt}
\begin{tabularx}{\textwidth}{@{}l|c|CCCC|CCCC@{}}
  \toprule
  \multirow{2}{*}{\textbf{Method}} &
  \multirow{2}{*}{\textbf{Prompt Format}} &
  \multicolumn{2}{c}{Individual Val} &
  \multicolumn{2}{c|}{Group Val} &
  \multicolumn{2}{c}{Individual Test} &
  \multicolumn{2}{c}{Group Test} \\
  \cmidrule(lr){3-4}
  \cmidrule(lr){5-6}
  \cmidrule(lr){7-8}
  \cmidrule(lr){9-10}
  & &
  MAE$\downarrow$ & RMSE$\downarrow$ &
  MAE$\downarrow$ & RMSE$\downarrow$ &
  MAE$\downarrow$ & RMSE$\downarrow$ &
  MAE$\downarrow$ & RMSE$\downarrow$ \\

  \midrule

  Qwen3-VL-8B~\shortcite{bai2025qwen3} & Text & 34.25 & 83.37 & 1.15 & 5.30 & 40.07 & 86.50 & 10.67 & 32.21 \\

  LLaVA-NeXT-Mistral-7B~\shortcite{liu2024llavanext} & Text & 48.59 & 92.51 & 3.21 & 7.17 & 55.27* & 99.42* & 11.85 & 34.51 \\
  
  InternVL3.5-8B~\shortcite{wang2025internvl3} & Text & 38.23 & 85.11 & 1.56 & 5.09 & 46.56 & 95.97 & 9.31 & 30.47 \\



  Qwen3.5-27B~\shortcite{qwen3.5} & Text & 40.00 & 91.39 & 0.87 & 5.33 & 38.28 & 82.26 & 5.84 & \underline{19.86} \\


  GLM-4.5V~\shortcite{hong2025glm} & Text & \underline{29.90} & \textbf{75.22} & 0.56 & 1.52 & 44.50 & 131.16 & 8.83 & 27.93 \\

  Qwen3.7-Max~\shortcite{qwen2026qwen37max} & Text & 36.66 & 90.50 & \textbf{0.38} & \textbf{1.15} & 31.02 & 84.96 & \underline{5.55} & 22.74 \\

  Claude Opus 5~\shortcite{anthropic2026claudeopus5} & Text & 30.09 & 82.05 & \underline{0.52} & \underline{1.47} & \underline{22.05} & \underline{49.16} & 10.26 & 33.76 \\

  GPT-5.6 Sol~\shortcite{openai2026gpt56} & Text & \textbf{27.22} & \underline{81.47} & 0.62 & 2.59 & \textbf{15.94} & \textbf{33.54} & \textbf{5.46} & \textbf{18.57} \\


  \hline
  \midrule

  CounTR~\shortcite{liu2022countr} & Visual &
  19.82 & 35.70 &
  4.01 & 10.51 &
  28.97 & 48.33 &
  9.16 & 18.98 \\

  DAVE~\shortcite{Pelhan_2024_CVPR} & Visual &
  15.91 & 29.67 &
  4.52 & 10.87 &
  13.83 & \textbf{22.07} &
  6.94 & 12.06 \\

  CACViT~\shortcite{Wang_Xiao_Cao_Lu_2024} & Visual &
  18.44 & 34.80 &
  1.51 & 2.67 &
  17.21 & 35.98 &
  4.46 & 12.97 \\

  CounTX~\shortcite{Amini-Naieni_2023_BMVC} & Text &
  20.28 & 37.05 &
  4.02 & 6.79 &
  25.01 & 61.94 &
  18.58 & 48.72 \\

  CLIP-Count~\shortcite{10.1145/3581783.3611789} & Text &
  22.22 & 36.11 &
  2.05 & 3.10 &
  38.17 & 82.22 &
  16.47 & 54.23 \\

  GroundingREC~\shortcite{Dai_2024_CVPR} & Text &
  \textbf{10.03} & \underline{26.89} &
  \underline{0.97} & \underline{2.02} &
  23.17 & 60.48 &
  \underline{3.59} & \underline{10.60} \\

  T2ICount~\shortcite{Qian_2025_CVPR} & Text &
  \underline{11.59} & \textbf{25.43} &
  12.71 & 53.06 &
  16.17 & 41.83 &
  23.53 & 68.03 \\

  CountSE~\shortcite{Liu_2025_ICCV} & Text &
  13.20 & 41.60 &
  \textbf{0.76} & \textbf{1.62} &
  14.70 & 29.18 &
  10.35 & 31.75 \\

  CountGD~\shortcite{NEURIPS2024_57c56985} & Text + Visual &
  13.05 & 34.05 &
  1.77 & 7.79 &
  \underline{13.54} & 38.68 &
  9.21 & 28.30 \\

  \midrule

  \textbf{GICount (Ours)} & Text + Visual &
  13.07 & 31.98 &
  2.18 & 9.31 &
  \textbf{12.20} & \underline{28.89} &
  \textbf{1.85} & \textbf{5.49} \\

  \bottomrule
\end{tabularx}
}
\caption{Main results on the individual- and group-level validation and test sets of BunchCount. The best and second-best results are shown in \textbf{bold} and \underline{underlined}, respectively. Results marked with * are computed after excluding samples with anomalous model output, details are provided in Appendix~B.2.}
\label{tab:main-results}
\end{table*}

\subsection{Training and Inference}

For each individual sample, we follow the objective design of CountGD and optimize a detection loss consisting of a localization term $\mathcal{L}_{\mathrm{loc}}$ and a classification term $\mathcal{L}_{\mathrm{cls}}$.

\noindent\textbf{Localization Loss.}
The localization loss measures the discrepancy between the predicted point coordinate $\hat{c}_i$ and its matched GT coordinate $c_i$ using the $\ell_1$ norm:
\begin{equation}
\mathcal{L}_{\mathrm{loc}}
=\sum_{i=1}^{N}\left\lVert \hat{c}_i-c_i\right\rVert_1,
\end{equation}
where $N$ is the number of predictions matched to GT targets.

\noindent\textbf{Classification Loss.}
The classification loss measures the discrepancy between predicted classes and their matched labels:
\begin{equation}
\mathcal{L}_{\mathrm{cls}}
=\operatorname{FocalLoss}(\hat{L},Y),
\end{equation}
where $\hat{L}$ denotes the predicted classification confidence and $Y$ denotes the class labels.
For each paired individual- and group-level sample, the overall objective combines the detection losses of the two samples with the relation loss:
\begin{equation}
\mathcal{L}
=\mathcal{L}_{\mathrm{det}}^{\mathrm{ind}}
+\lambda_2\mathcal{L}_{\mathrm{det}}^{\mathrm{grp}}
+\lambda_3\mathcal{L}_{\mathrm{rel}},
\end{equation}
\begin{equation}
\mathcal{L}_{\mathrm{det}}
=\mathcal{L}_{\mathrm{loc}}
+\lambda_1\mathcal{L}_{\mathrm{cls}}.
\end{equation}
Here, $\lambda_1$, $\lambda_2$, and $\lambda_3$ balance the classification, group-level detection, and relation objectives, respectively.

During training, the individual- and group-level samples associated with the same image are paired and jointly fed into the model, enabling relation aggregation and computation of the relation loss. At inference time, each sample is processed independently through the underlying detection pipeline, which predicts the locations of all queried targets and obtains the count from the number of predictions.

\section{Experiments}

\subsection{Experimental Setup}

\noindent\textbf{Implementation Details.}
We adopt GroundingDINO-B as the backbone and initialize it with the publicly released CountGD checkpoint, while randomly initializing the proposed relation-modeling module. The model is then trained on BunchCount for 30 epochs. During training, the Swin Transformer image encoder and the BERT text encoder are frozen, and all remaining trainable parameters are optimized with Adam. The initial learning rate is set to $1\times10^{-4}$ and multiplied by $0.1$ every 10 epochs. Training is conducted on a single NVIDIA RTX 5090 GPU with a batch size of 4 and takes approximately 10 hours. We set the numbers of sampled positive and negative relation sets, $N^{+}$ and $N^{-}$, to 2, the maximum relation-set size $M$ to 5, and the loss weights $\lambda_2$ and $\lambda_3$ to 1. Due to space constraints, complete implementation details are provided in Appendix~B.1.

\noindent\textbf{Evaluation Metrics.}
Each test image in BunchCount gives rise to one individual-level counting sample and one group-level counting sample. Since the two counting units differ substantially in numerical scale, we report their performance separately. Following standard practice in visual counting, we evaluate all methods using Mean Absolute Error (MAE) and Root Mean Squared Error (RMSE), with detailed definitions provided in Appendix~C.

\subsection{Results on BunchCount} 

\begin{table}[!t]
\centering
{\small
\setlength{\tabcolsep}{2.5pt}
\begin{tabularx}{\columnwidth}{@{}l|CCCC@{}}
  \toprule
  \multirow{2}{*}{\textbf{Method}} &
  \multicolumn{2}{c}{Individual Test} &
  \multicolumn{2}{c}{Group Test} \\
  \cmidrule(lr){2-3}\cmidrule(lr){4-5}
  & MAE$\downarrow$ & RMSE$\downarrow$
  & MAE$\downarrow$ & RMSE$\downarrow$ \\
  \midrule
  w/o relation modeling & 13.54 & 38.68 & 9.21 & 28.30 \\
  w/o counting-unit prompt & 18.36 & 57.39 & 1.93 & 6.79 \\
  full~(Ours) & \textbf{12.20} & \textbf{28.89} & \textbf{1.85} & \textbf{5.49} \\
  \bottomrule
\end{tabularx}
}
\caption{Effect of relation modeling and counting-unit prompt on GICount framework.}
\label{tab:ablation-results}
\end{table}

\begin{table}[!t]
\centering
{\small
\setlength{\tabcolsep}{2.5pt}
\begin{tabularx}{\columnwidth}{@{}l|c|CCCC@{}}
  \toprule
  \multirow{2}{*}{\textbf{Method}} &
  \multirow{2}{*}{\textbf{Train Setting}} &
  \multicolumn{2}{c}{Individual Test} &
  \multicolumn{2}{c}{Group Test} \\
  \cmidrule(lr){3-4}\cmidrule(lr){5-6}
  & & MAE$\downarrow$ & RMSE$\downarrow$
  & MAE$\downarrow$ & RMSE$\downarrow$ \\
  \midrule
  \multirow{3}{*}{CountGD}
  & Ind.-only & 10.92 & 26.01 & 61.73 & 100.66 \\
  & Grp.-only & 73.95 & 115.76 & 1.38 & 3.98 \\
  & Joint & 13.54 & 38.68 & 9.21 & 28.30 \\
  \midrule
  GICount~(Ours) & Joint & \textbf{12.20} & \textbf{28.89} & \textbf{1.85} & \textbf{5.49} \\
  \bottomrule
\end{tabularx}
}
\caption{The comparison results under different training settings on the individual- and group-level test sets.}
\label{tab:counting-unit}
\end{table}

\begin{figure*}[!t]
  \centering
  \includegraphics[width=\textwidth]{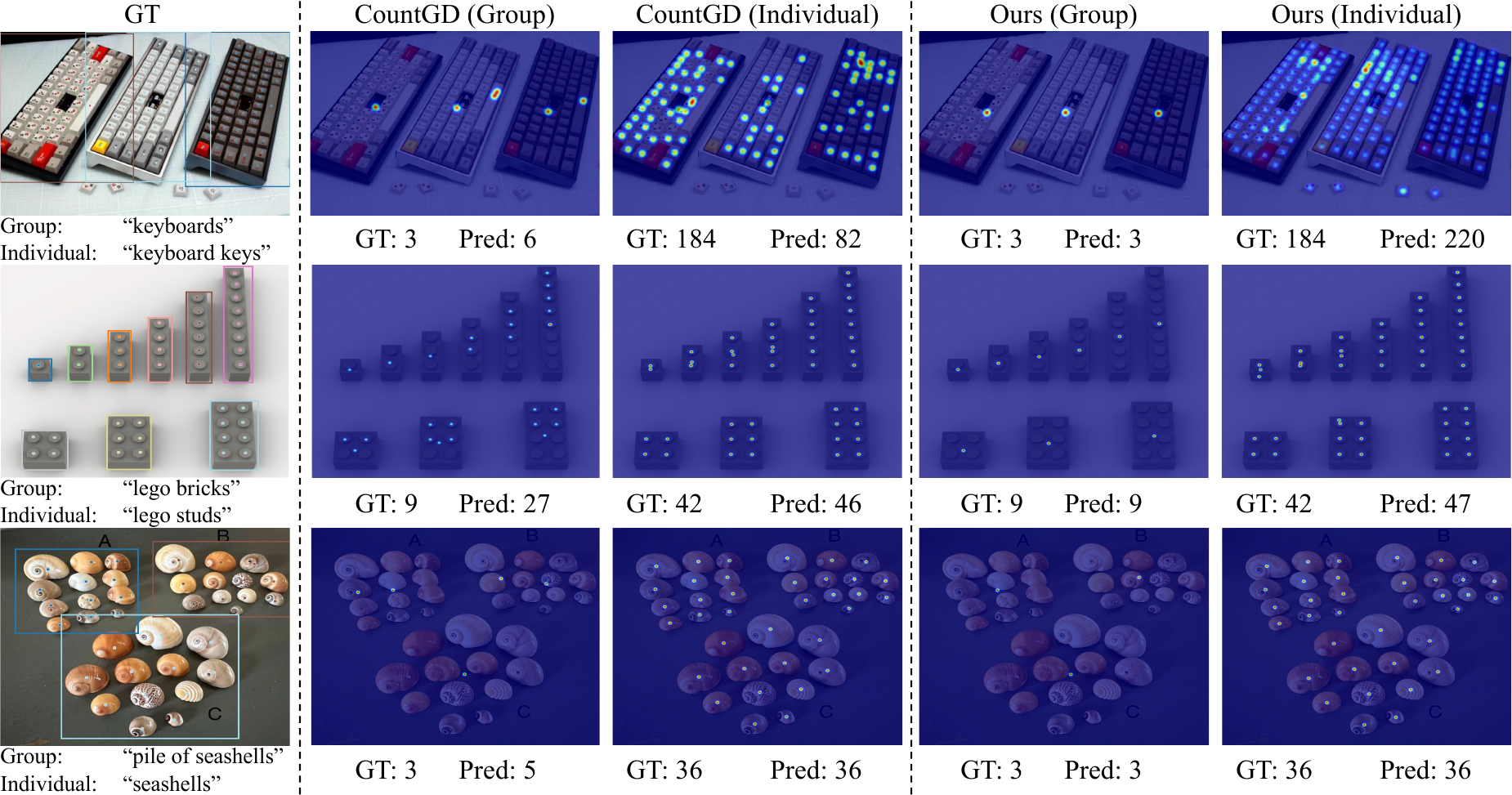}
  \caption{Qualitative results of CountGD and GICount on BunchCount under paired group- and individual-level prompts. }
  \label{fig:qualitative}
\end{figure*}

We compare against several representative class-agnostic counting methods spanning visual-exemplar, text, and multimodal prompting, and additionally report results from representative vision-language models (VLMs) in Table~\ref{tab:main-results}. 
The counting methods are fine-tuned on BunchCount by uniformly sampling from the merged individual- and group-level training samples, with implementation, evaluation details and cost analysis provided in Appendix~B.2, B.3 and D.
CountGD achieves the lowest individual-level MAE among the counting baselines (13.54), but records a group-level MAE of 9.21. In contrast, GroundingREC obtains the strongest baseline group-level result~(3.59 MAE) despite a considerably higher individual-level MAE of 23.17, possibly because its fine-grained attribute partially align with the structural semantics in group prompts. This ranking reversal shows that conventional instance-counting performance does not reliably predict the ability to count semantic groups, highlighting the distinct capability evaluated by BunchCount.

GICount achieves 1.85 and 5.49 of MAE and RMSE on the group test set, reducing the errors of GroundingREC by 48.5\% and 48.2\%. It also obtains an individual-level MAE of 12.20 and an RMSE of 28.89. Overall, these results establish GICount as a strong unified baseline that substantially advances semantic-group counting while retaining strong individual-level counting performance. As illustrated in Figure~\ref{fig:qualitative}, GICount places one prediction on each intended group and shifts its predictions to the constituent instances when given the corresponding individual-level prompt.

\subsection{Ablation Study on GICount}

We ablate the two key components of GICount in Table~\ref{tab:ablation-results}. 
w/o relation modeling uses only individual- and group-level detection losses, while w/o counting-unit prompt applies relation supervision with the mean-pooled full group prompt as the anchor.
Adding relation modeling sharply reduces group-level MAE from 9.21 to 1.93, showing that containment-based supervision is crucial for semantic-group counting. 
However, using the full prompt as the relation anchor entangles category and counting-unit semantics, increasing individual-level MAE from 13.54 to 18.36. 
By isolating group-structure semantics as the anchor, the full model preserves the group-level gain and restores individual counting, achieving 1.85 group MAE and 12.20 individual MAE. 
These results show that relation modeling enhances group counting, and counting-unit prompts are necessary to avoid cross-granularity interference.

\subsection{Counting-units Analysis}

We examine whether mixing individual- and group-level samples is sufficient to learn a unified counter in Table~\ref{tab:counting-unit}. We train three CountGD variants using individual samples only, group samples only, or uniformly mixed samples, and report GICount under the same joint training setting.
The two single-unit models show clear specialization. Individual-only and group-only achieve 10.92 and 1.38 MAE on their respective units, while their cross-unit errors reach 61.73 and 73.95. Since BunchCount pairs individual and group queries for the same image and category, this reversal shows that category identity alone does not specify the intended counting granularity and establishes the counting unit as an essential conditioning factor. Directly mixing both units enables a single CountGD model to process both query types, but its group MAE increases from 1.38 under group-only training to 9.21, exposing semantic granularity conflict in joint training.

\section{Conclusions}

In this paper, we present GIC, a new counting task that requires models to count either individuals or the semantic groups they form according to the specified counting unit. Furthermore, we construct BunchCount, a real-world benchmark with paired individual-group annotations and explicit containment relations, and propose GICount, a counting-unit guided relational framework. We perform systematic experiments and establish GICount as a strong baseline for this task. This work moves visual counting beyond individual instances and provides a foundation for semantic unit-aware counting in practical scenarios such as retail inventory, agriculture, warehousing, and logistics.

\bibliography{aaai2027}

\clearpage
\appendix 

\section*{Appendix}

\section{Additional Details of BunchCount}

\subsection{Semantic Groups and Annotation Scope}

A group is primarily a semantic and perceptual unit rather than one determined by a single geometric rule. In common visual understanding, multiple objects of the same category can be perceived as a coherent and countable whole through cues such as regular arrangement, physical connection, fixed cardinality, spatial aggregation, or shared containment. In some cases, this perception also coincides with a part--whole organization, where repeated objects jointly form a familiar whole, such as beads forming a bracelet or lenses forming sunglasses. Although these groups differ substantially in appearance and organization, they share the same counting interpretation: the visible objects can be counted either individually or according to the higher-level groups they form.

To reduce ambiguity and ensure reliable annotations, BunchCount adopts a conservative image-selection protocol. We retain scenes in which the constituent objects are individually identifiable, the perceived groups have relatively clear boundaries, and group membership can be determined with reasonable consistency. Such boundaries may be supported by visible containers or physical connections, regular spatial organization, or clear separation between neighboring groups, for example, when the distance between groups is noticeably larger than the spacing among objects within each group. Cases with highly ambiguous group existence, boundaries, or constituent membership are excluded. For each retained image, we annotate individual objects with center points, semantic groups with bounding boxes, and explicitly associate each group with its constituent individuals, while allowing some individuals to remain ungrouped.

\subsection{Taxonomy of Group Structures}

To characterize the diverse ways in which individual objects form semantic groups, we organize the groups in BunchCount into four structure types according to their most representative organizing characteristics: structural arrangement, physical connection, fixed cardinality, and spatial aggregation. A group may exhibit multiple perceptual cues, while its assigned type captures the cue that most directly supports its interpretation as a coherent and countable unit. This assignment provides each group with a single structural label and supports consistent dataset annotation and analysis.

\noindent\textbf{Structural arrangement.}
This type describes groups formed through a regular or ordered spatial configuration among their constituent objects. The relative positions and ordering of the objects provide the primary cue for perceiving them as a coherent unit. Representative forms include stacks, rows, and columns, such as a stack of sugar cubes or a row of books. The same structure can appear across object categories with substantially different appearances, while preserving a recognizable organization.

\noindent\textbf{Physical connection.}
This type describes groups whose constituent objects form a unified whole through visible attachment, linkage, or binding. The physical relation among the objects provides the primary cue for identifying the group and its members. Typical examples include grapes connected as a bunch, objects tied into a bundle, and beads linked into a string or bracelet. These groups often have flexible shapes, while their connections maintain a stable group identity.

\noindent\textbf{Fixed cardinality.}
This type describes groups defined primarily by a conventional or functional number of constituent objects. The prescribed cardinality determines how the visible objects are combined into countable units across different spatial configurations. Representative examples include a pair of shoes and two lenses forming a pair of sunglasses. The objects may vary in position, orientation, and appearance, while their cardinal relation provides a consistent counting interpretation.

\noindent\textbf{Spatial aggregation.}
This type describes groups formed through local spatial clustering or a shared container or supporting region. Spatial coherence provides the primary cue for associating the constituent objects with the same countable unit. Representative forms include piles and collections placed in boxes, bowls, bags, or plates, such as a pile of candies or a box of strawberries. Shared containment is included in this type because the containing region provides a clear spatial organization and group boundary.

\subsection{Data Collection and Image Selection}

We first identified object categories that commonly form countable groups. Candidate images were collected from FSC-147, Locount, COCO, OpenImages, and the web. We manually inspected images from the existing datasets. For web collection, we searched using either category names or combinations of group-structure terms and category names, such as ``stack of plates,'' ``pair of shoes,'' and ``pile of stones.'' All images are licensed for non-commercial use, and the released BunchCount dataset includes the original image files. \textbf{The dataset and annotations will be made publicly available upon publication}.

We selected real-world images containing multiple objects of the same category and clearly distinguishable groups, allowing both individual- and group-level annotation. We excluded low-resolution images, images with severe occlusion or object density that prevented individual identification, and images in which group boundaries or member assignments were ambiguous. Synthetic and AI-generated images were also excluded.

\subsection{Data Annotation}

For each image, annotators first specified the individual-object category and the corresponding group description. All visible individual objects were annotated with center points, while each semantic group was annotated with a tight bounding box covering its visible constituent objects. Each individual point was then explicitly associated with at most one group, and individuals not belonging to any group were retained as ungrouped instances. 

Groups without an explicit enclosing boundary were required to contain at least two visible individuals. A group with only one visible constituent was retained only when its identity was independently supported by a clear boundary, such as a shared container, or by a recognizable whole-object structure whose remaining constituents were occluded or outside the image. For fixed-cardinality groups, membership additionally followed the corresponding cardinal constraint. To support exemplar-based counting, three individual instances and one to three semantic groups were randomly selected in each image. Tight instance boxes were annotated for the selected individual exemplars, while the corresponding group annotations were used as group exemplars.

The annotations were produced by six annotators using our custom annotation platform and subsequently cross-checked for textual consistency, duplicated or missing point and box annotations, and incorrect group--individual associations. Samples whose group description, boundary, or membership remained ambiguous during cross-checking were removed rather than resolved through an arbitrary assignment. This process filtered out 58 candidate images, resulting in the final set of 1,330 images.

\subsection{Dataset Split and Statistics}

We split BunchCount by target category into mutually disjoint training, validation, and test sets. For categories overlapping with FSC-147, we follow its original split to maintain compatibility. The resulting splits contain 63, 9, and 14 categories with 869, 151, and 310 images, respectively.

\begin{figure*}[!t]
  \centering
  \includegraphics[width=\textwidth]{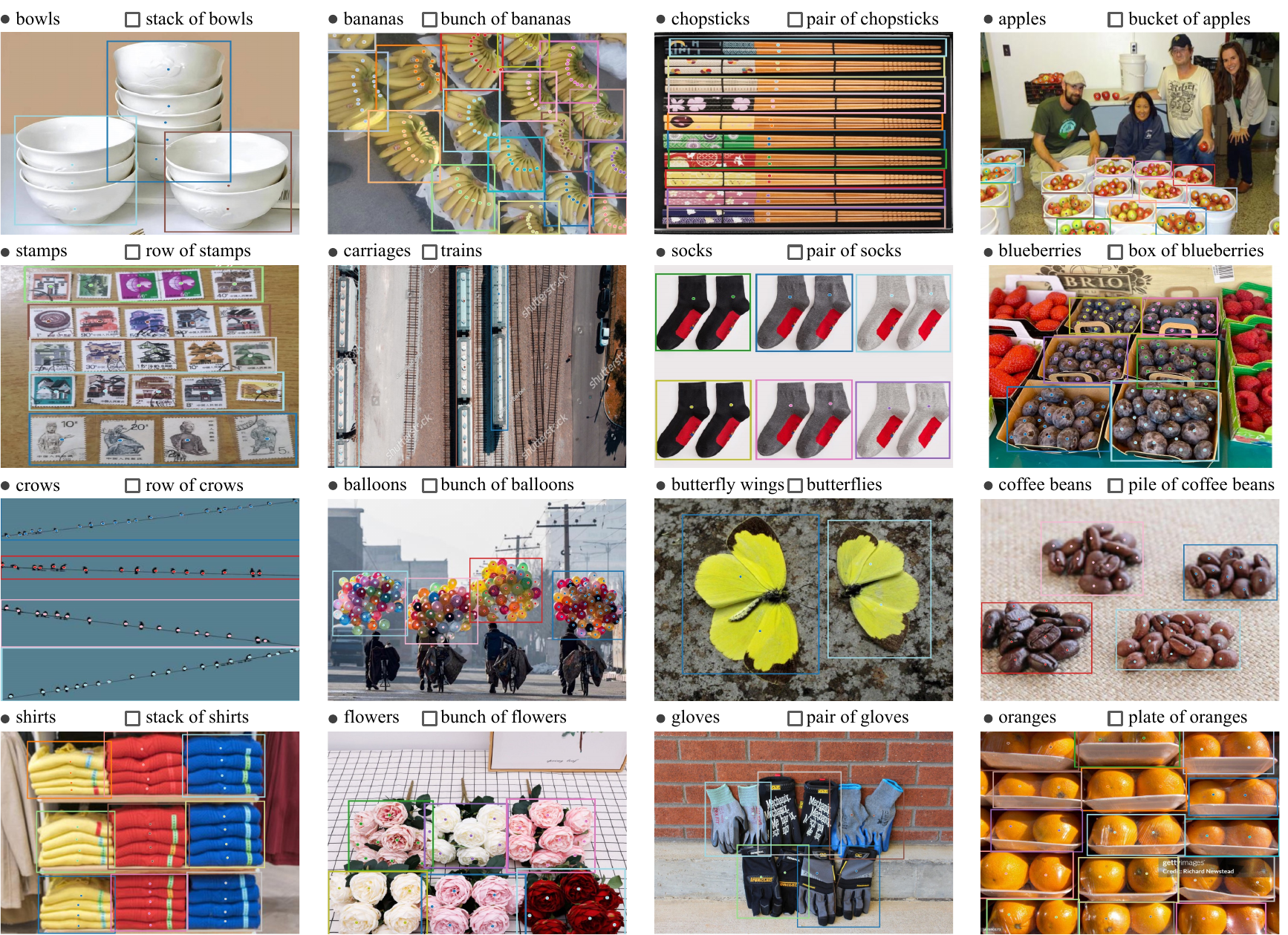}
  \caption{Additional samples from BunchCount.}
  \label{fig:additional_samples}
\end{figure*}

Group-structure descriptions are allowed to recur across splits. Unlike object categories, which cover diverse appearances and semantics, semantic groups are formed through a limited set of recurring structures, including arrangement, physical connection, fixed cardinality, and spatial aggregation. This design evaluates whether group-structure semantics learned from training categories can transfer to unseen object categories. The average annotation counts and the numbers of samples for the four group-structure types in each split are summarized in Tables~\ref{tab:split-statistics} and \ref{tab:split-structures}. Additional samples from BunchCount are shown in Figure~\ref{fig:additional_samples}.

\begin{table}[!t]
\centering
{\small
\setlength{\tabcolsep}{4pt}
\begin{tabularx}{\columnwidth}{@{}l|CCCC@{}}
  \toprule
  \textbf{Split} &
  \textbf{Classes} &
  \textbf{Images} &
  \textbf{Ind. avg.} &
  \textbf{Grp. avg.} \\
  \midrule
  Train & 63 & 869 & 58 & 6  \\
  Val   & 9  & 151 & 69 & 5  \\
  Test  & 14 & 310 & 90 & 16 \\
  \bottomrule
\end{tabularx}
}
\caption{Split-wise statistics of BunchCount.}
\label{tab:split-statistics}
\end{table}

\begin{table}[!t]
\centering
{\small
\setlength{\tabcolsep}{3pt}
\begin{tabularx}{\columnwidth}{@{}l|c|CCCC@{}}
  \toprule
  \multirow{2}{*}{\textbf{Split}} &
  \multirow{2}{*}{\textbf{Images}} &
  \multicolumn{4}{c}{\textbf{Group Structure}} \\
  \cmidrule(lr){3-6}
  & &
  \shortstack{Physical\\connection} &
  \shortstack{Spatial\\aggregation} &
  \shortstack{Fixed\\cardinality} &
  \shortstack{Structural\\arrangement} \\
  \midrule
  Train & 869 & 194 & 237 & 98 & 340 \\
  Val   & 151 & 47  & 42  & 6  & 56  \\
  Test  & 310 & 39  & 171 & 45 & 55  \\
  \bottomrule
\end{tabularx}
}
\caption{Distribution of group-structure types across the dataset splits.}
\label{tab:split-structures}
\end{table}



\section{Additional Implementation Details}

\subsection{Additional GICount Details}

\noindent\textbf{Architecture.} GICount is built on CountGD and retains its detection backbone, RoI-based exemplar injection, and detection head. For counting-unit prompting, predefined prompt annotations separate the group descriptor from the object category. The BERT-encoded group descriptor is used as the semantic anchor; for holistic group names without an explicit descriptor, two learnable counting-unit embeddings are inserted before the feature enhancer. Relation sets are encoded by an independent two-layer Transformer encoder with a hidden dimension of 256, eight attention heads, an FFN dimension of 2048, GELU activation, 0.1 dropout, pre-layer normalization, and a final LayerNorm. Each set contains at most five individual queries preceded by a learnable \texttt{[REL]} token, whose output is used as the relation representation.

\noindent\textbf{Training.} GICount is initialized from the publicly released CountGD checkpoint, while the relation encoder, [REL] token, and learnable counting-unit embeddings are randomly initialized. Each image forms paired individual- and group-level samples, which are processed through two forward passes of the shared model. For group-level detection supervision, each annotated group box is represented by its center point, while the original box is retained as a visual exemplar. CountGD takes three exemplars per target; when fewer than three group exemplars are available, the last exemplar is repeated for padding. For relation supervision, queries from the final decoder layer of the individual-level pass are matched to the ground-truth points and organized using the annotated group--individual relations. We sample two positive sets from individuals within the same group and two negative sets from different groups, with at most five queries per set. As described in the main text, the relation loss combines a triplet loss and a pairwise positive-alignment loss. We search their relative weights over \{0.1:0.9,0.3:0.7,0.5:0.5,0.7:0.3,0.9:0.1\} and set both weights to 0.5~(1:1). The triplet margin is set to 0.2. The individual- and group-level detection losses are also equally weighted. We freeze the Swin image encoder and BERT text encoder and optimize all remaining parameters using Adam with a learning rate of $1\times10^{-4}$, weight decay of $1\times10^{-4}$, and global gradient clipping of 0.1. The learning rate is reduced by a factor of ten every ten epochs. The maximum number of sampled label phrases is set to 40. The model is trained for 30 epochs with a batch size of 4 on one NVIDIA RTX 5090 GPU, taking approximately 10 hours. The relation module is disabled during inference, which follows the original CountGD localization and confidence-threshold counting procedure.

\subsection{VLM Evaluation}

We evaluate all vision-language models without finetuning on BunchCount. Each model receives an image and a text prompt specifying either the individual or group counting target. We disable reasoning mode and set the temperature to zero when supported, with the maximum output length set to 32 tokens.

We designed three prompt formats---\textit{direct}, \textit{exclude}, and \textit{detailed}---with progressively increasing instruction specificity, evaluated them using Qwen3-VL-8B, and selected the second format for all VLM evaluations. The individual- and group-level prompts are defined as follows:

\noindent\textbf{Individual prompt:} \textit{Count the visible individual instances matching ``\{target\_text\}'' in this image. Count individual items only, not groups or clusters of items. Return only one non-negative integer and no other text.}

\noindent\textbf{Group prompt:} \textit{Count the visible groups matching ``\{target\_text\}'' in this image. Count groups only, not the individual items inside the groups. Return only one non-negative integer and no other text.}

We extract a numerical count from each model response whenever possible. Responses from which no valid number can be parsed are treated as invalid and excluded from evaluation. Across all evaluated models, only LLaVA-NeXT-Mistral-7B produced such invalid responses, resulting in the exclusion of six samples.

\subsection{Adaptation of Counting Methods}

We use the official implementations and publicly released checkpoints for all counting baselines, and continue training on BunchCount following their official training recipes. Each image yields one individual-level sample and one group-level sample, which are randomly mixed with equal sampling probability. Individuals are represented by their annotated points, while group bounding boxes are converted to center points. For methods requiring three visual exemplars, available group exemplars are duplicated when necessary. All other settings follow the official implementations.

For GroundingREC, we preserve its paired referring-expression training instead of treating the two counting levels as independent samples. Each image is associated with an individual-level and a group-level attribute expression, and the original same-image contrastive objective is retained, such that the two counting-unit expressions serve as negatives to each other.

\section{Additional Evaluation Metrics}

Following standard practice in visual counting, we evaluate counting accuracy using Mean Absolute Error (MAE) and Root Mean Squared Error (RMSE):
\begin{equation}
\mathrm{MAE}=\frac{1}{N}\sum_{i=1}^{N}\left|\hat{y}_i-y_i\right|,\
\mathrm{RMSE}=\sqrt{\frac{1}{N}\sum_{i=1}^{N}\left(\hat{y}_i-y_i\right)^2},
\end{equation}
where $N$ is the number of evaluation samples, and $\hat{y}_i$ and $y_i$ denote the predicted and ground-truth counts of the $i$-th sample, respectively. We compute these metrics separately for individual- and group-level samples.

For model selection, we use an average normalized MAE to balance performance across the two counting units. Specifically, we evaluate the candidate model weights on the validation set and normalize the MAE of each counting unit by the lowest corresponding MAE among all candidates:
\begin{equation}
\mathrm{NMAE}_{\mathrm{avg}}
=
\frac{1}{2}
\left(
\frac{\mathrm{MAE}_{\mathrm{ind}}}{\mathrm{MAE}_{\mathrm{ind}}^{\min}}
+
\frac{\mathrm{MAE}_{\mathrm{grp}}}{\mathrm{MAE}_{\mathrm{grp}}^{\min}}
\right),
\end{equation}
where $\mathrm{MAE}_{\mathrm{ind}}^{\min}$ and $\mathrm{MAE}_{\mathrm{grp}}^{\min}$ denote the lowest individual- and group-level MAEs among the candidate weights, respectively. We select the weight with the lowest $\mathrm{NMAE}_{\mathrm{avg}}$, while reporting the original MAE and RMSE for final evaluation.

\section{Computational Cost}

\begin{table}[!t]
\centering
{\small
\setlength{\tabcolsep}{3pt}
\begin{tabularx}{\columnwidth}{@{}l|c|C|c@{}}
  \toprule
  \textbf{Method} &
  \shortstack{\textbf{Parameters}} &
  \textbf{Input} &
  \shortstack{\textbf{FLOPs} \textbf{per Forward}} \\
  \midrule
  CounTR
  & 99.69M
  & $384^2$
  & 168.66G \\

  CLIP-Count
  & 166.09M
  & $384^2$
  & 54.41G \\

  GroundingREC
  & 173.10M
  & $800\times1333$
  & 1.621T \\

  CountGD
  & 237.31M
  & $800{\times}1333$
  & 1.368T \\

  GICount~(Ours)
  & 239.94M
  & $800{\times}1333$
  & 1.375T / 1.377T \\
  \bottomrule
\end{tabularx}
}
\caption{Computational cost of representative counting methods under their reference input configurations. The two FLOPs values of GICount correspond to individual- and group-level queries, respectively.}
\label{tab:computational-cost}
\end{table}

Table~\ref{tab:computational-cost} compares the model size and per-sample computation of representative counting methods. GICount contains 239.94M parameters, only 2.63M, or 1.1\%, more than its CountGD backbone. The group-individual relation module is used only during training and therefore introduces no additional relation-modeling branch at inference. Under an $800\times1333$ input, GICount requires 1.375 and 1.377 TFLOPs for individual- and group-level queries, respectively, indicating nearly identical inference costs across the two counting units. Training the full model for 30 epochs takes approximately 10 hours on a single NVIDIA RTX 5090 GPU.

\section{Results across Group Structures}

\begin{table}[!t]
\centering
{\small
\setlength{\tabcolsep}{2.5pt}
\begin{tabularx}{\columnwidth}{@{}l|C|C|C|C@{}}
  \toprule
  \textbf{Method} &
  \shortstack{Physical\\Connection\\(39)} &
  \shortstack{Spatial\\Aggregation\\(171)} &
  \shortstack{Fixed\\Cardinality\\(45)} &
  \shortstack{Structural\\Arrangement\\(55)} \\
  \midrule
  CountGD
  & 1.64 / 4.11
  & 0.64 / 1.98
  & 58.22 / 74.05
  & 1.11 / 1.93 \\

  GICount~(Ours)
  & 1.18 / 1.95
  & 0.74 / 2.24
  & 8.11 / 13.55
  & 0.62 / 1.18 \\
  \bottomrule
\end{tabularx}
}
\caption{Group-level counting results across different group structures on the BunchCount test set. Each entry reports MAE / RMSE, and the number of samples in each subset is shown in parentheses.}
\label{tab:group-structure-results}
\end{table}

We further break down the group-level results according to the four group structures in BunchCount, as reported in Table~\ref{tab:group-structure-results}. CountGD and GICount produce similar errors on spatial aggregation, whereas GICount records lower errors on physical connection and structural arrangement. Fixed cardinality remains the most difficult subset for both models.

\subsection{Additional Ablation Study}

\begin{table}[!t]
\centering
{\small
\setlength{\tabcolsep}{2.5pt}
\begin{tabularx}{\columnwidth}{@{}l|CCCC@{}}
  \toprule
  \multirow{2}{*}{\textbf{Method}} &
  \multicolumn{2}{c}{Individual Test} &
  \multicolumn{2}{c}{Group Test} \\
  \cmidrule(lr){2-3}\cmidrule(lr){4-5}
  & MAE$\downarrow$ & RMSE$\downarrow$
  & MAE$\downarrow$ & RMSE$\downarrow$ \\
  \midrule
  mean-pooling aggregator
  & 16.02 & 44.13 & 2.55 & 8.62 \\
  group-query anchor
  & 21.63 & 69.05 & 1.51 & 4.63 \\
  triplet loss only
  & 20.37 & 64.33 & 1.63 & 4.49 \\
  positive-alignment loss only
  & 18.80 & 57.10 & 1.85 & 5.41 \\
  full~(Ours)
  & \textbf{12.20} & \textbf{28.89}
  & 1.85 & 5.49 \\
  \bottomrule
\end{tabularx}
}
\caption{Additional ablations on the components of relation supervision.}
\label{tab:additional-ablation}
\end{table}

We further ablate the aggregation, anchor, and loss designs within relation supervision. All variants follow the same training and evaluation settings as the full model, with only the specified component modified. For relation aggregation, we replace the Transformer-based relation aggregator with mean pooling over the sampled individual-query features. For the relation anchor, we replace the group-structure representation extracted from the counting-unit prompt with the group-query representation. For the loss objective, we separately retain only the triplet term or only the pairwise positive-alignment term. Results are shown in Table~\ref{tab:additional-ablation}

\begin{figure*}[!t]
  \centering
  \includegraphics[width=\textwidth]{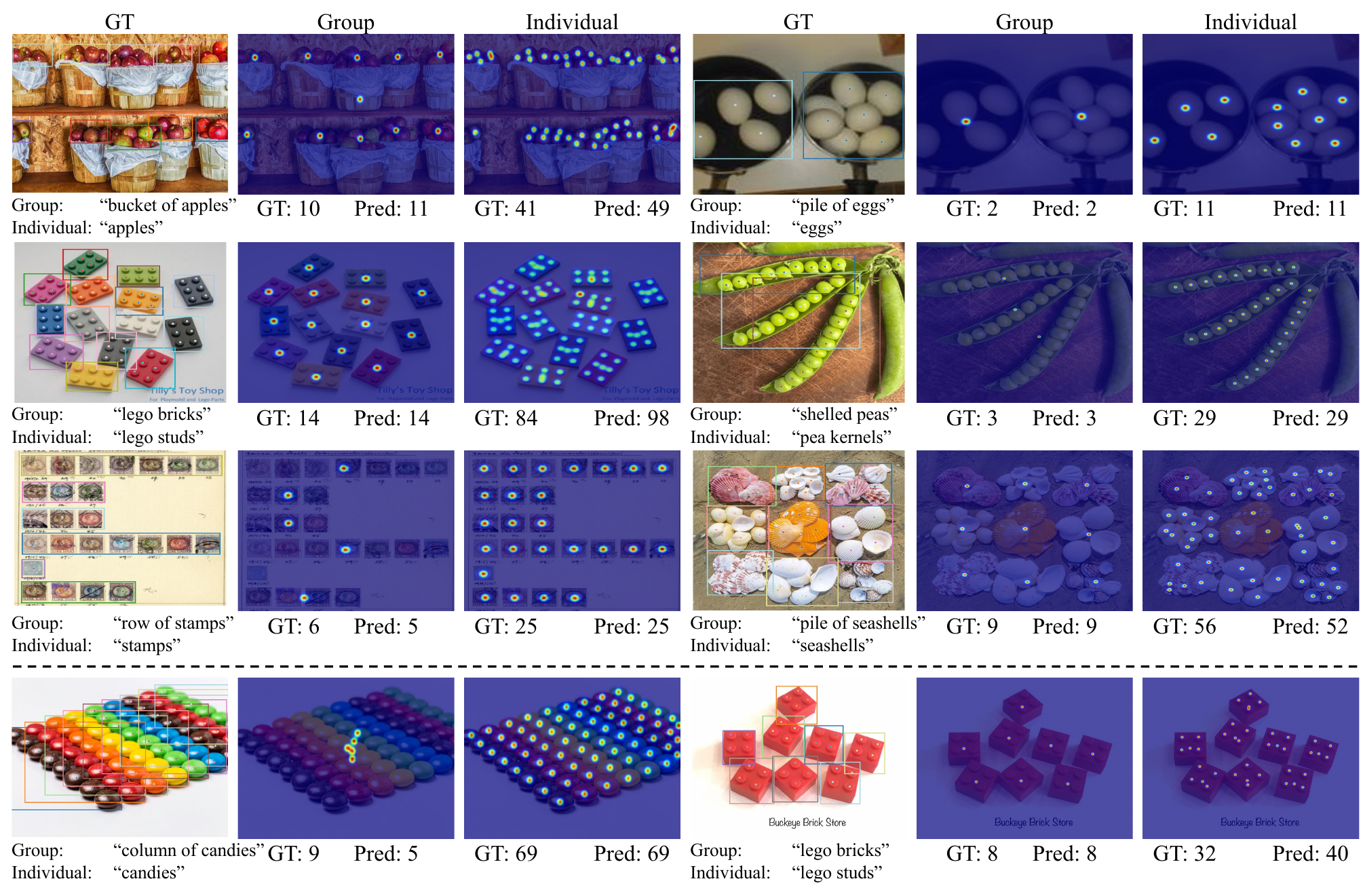}
  \caption{Additional qualitative results of GICount on BunchCount.}
  \label{fig:additional_qualitative}
\end{figure*}

\section{Additional Qualitative Results}

Figure~\ref{fig:additional_qualitative} presents additional qualitative results of GICount on diverse object categories and group structures. For each example, the ground-truth group annotations are shown together with the predicted group- and individual-level response maps under the corresponding counting-unit prompts. The examples above the dashed line illustrate representative predictions, while those below show several failure cases.

\end{document}